\def\year{2022}\relax
\documentclass[letterpaper]{article} 
\usepackage{aaai22}  
\usepackage{times}  
\usepackage{helvet}  
\usepackage{courier}  
\usepackage[hyphens]{url}  
\usepackage{graphicx} 
\usepackage{natbib}  
\usepackage{caption} 
\DeclareCaptionStyle{ruled}{labelfont=normalfont,labelsep=colon,strut=off} 
\usepackage{algorithm}
\usepackage{algorithmic}

\usepackage{newfloat}
\usepackage{listings}
\floatstyle{ruled}
\newfloat{listing}{tb}{lst}{}
\floatname{listing}{Listing}
\usepackage{amsmath,amssymb,amsfonts}
\usepackage{algorithmic}
\usepackage{graphicx}
\usepackage{textcomp}
\usepackage{xcolor}
\usepackage{booktabs}
\usepackage{multirow}
\usepackage{subcaption}

\title{Memory-augmented Adversarial Autoencoders for Multivariate Time-series Anomaly Detection with Deep Reconstruction and Prediction}
\author {
    Qinfeng Xiao,
    Shikuan Shao,
    Jing Wang 
}
\affiliations {
    Beijing Jiaotong University	\\
    {qfxiao, skshao, wj}@bjtu.edu.cn
}

\usepackage{bibentry}

\begin{document}

\maketitle

\begin{abstract}

Detecting anomalies for multivariate time-series without manual supervision continues a challenging problem due to the increased scale of dimensions and complexity of today's IT monitoring systems. Recent progress of unsupervised time-series anomaly detection mainly use deep autoencoders to solve this problem, i.e. training on normal samples and producing significant reconstruction error on abnormal inputs. However, in practice, autoencoders can reconstruct anomalies so well, due to powerful capabilites of neural networks. Besides, these approaches can be ineffective for identifying non-point anomalies, e.g. contextual anomalies and collective anomalies, since they solely utilze a point-wise reconstruction objective. To tackle the above issues, we propose MemAAE (\textit{Memory-augmented Adversarial Autoencoders with Deep Reconstruction and Prediction}), a novel unsupervised anomaly detection method for time-series. By jointly training two complementary proxy tasks, reconstruction and prediction, with a shared network architecture, we show that detecting anomalies via multiple tasks obtains superior performance rather than single-task training. Additionally, a compressive memory module is introduced to preserve normal patterns, avoiding unexpected generalization on abnormal inputs. Through extensive experiments, MemAAE achieves an overall F1 score of 0.90 on four public datasets, significantly outperforming the best baseline by 0.02. 
\end{abstract}

\section{Introduction}
Detecting anomalies for time-series is critical for various fields, e.g. web service maintenance \cite{xu2018unsupervised}, industry device monitoring \cite{su2019robust,10.1145/3357384.3357816}, cyber intrusion detection \cite{li2019mad} and anomalous rhythm detection \cite{zhou2019beatgan}. Since anomalies appeared in time-series indicate failures of machines or symptoms of human bodies, automatic anomaly detection is of significant importance for many practical applications. Traditional strategies detect anomalies via supervised classifiers \cite{10.1145/2815675.2815679}, namely considering the anomaly detection problem as a binary classification task. However, supervised anomaly detection faces exact challenges including imbalanced classes, heavy labeling cost and presence of unseen anomalous patterns, which limit the use of them in practical situations.  Recently, deep autoencoders have drawn extensive attention for time-series anomaly detection \cite{xu2018unsupervised,su2019robust,li2019mad,audibert2020usad}. By training variational autoencoders (VAE) or Generative Adversarial Networks (GAN) on normal data, the model is expected to produce higher reconstruction errors for anomalies. However, the problem of unsupervised multivariate time-series anomaly detection is still challenging due to the following questions: 


\begin{figure}[h]
\centering
\begin{subfigure}{0.48\textwidth}
	\centering
	\includegraphics[width=\linewidth]{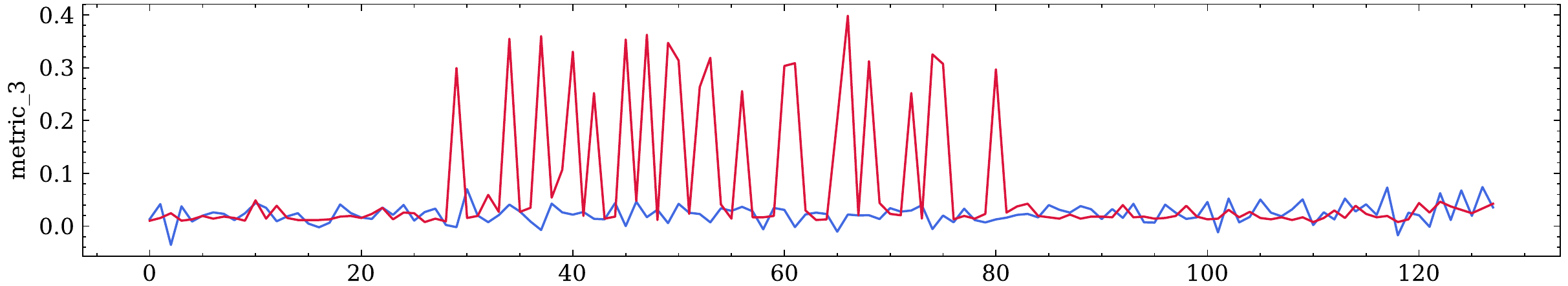}
	\caption{Point anomalies captured by autoencoders.}
	\label{fig:intro1}
\end{subfigure}
\begin{subfigure}{0.482\textwidth}
	\centering
	\includegraphics[width=\linewidth]{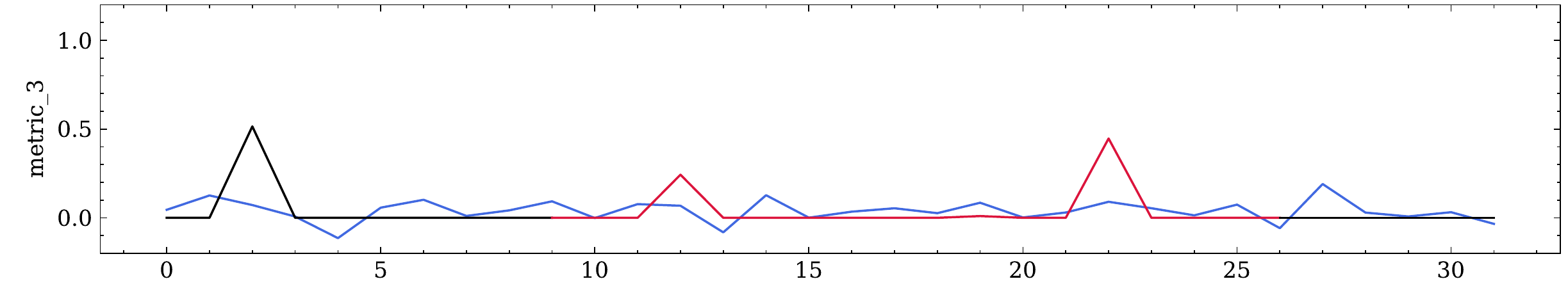}
	\caption{Contextual anomalies ignored by autoencoders.}
	\label{fig:intro2}
\end{subfigure}
\caption{Examples of anomaly detection using conventional reconstruction-based method, where black lines, red lines and blue lines indicate normal observations, anomalies and reconstructions respectively.}
\label{fig:series}
\end{figure}

1) \textit{Complexity of multivariate time-series.} Today’s developing of IT monitoring systems and increasing dimensions of time-series render the difficulty of anomaly detection. As internal complexities of data grow exponentially, autoencoders' capability of modeling data distribution has seriously overstretched. To this end, integration of modern techniques, e.g. GAN \cite{goodfellow2014generative}, are desired.

2) \textit{Heterogeneous types of anomalies.} As described in previous researches \cite{chandola2009anomaly,chalapathy2019deep}, anomalies are categorized into different classes, e.g. point anomaly, contextual anomaly and collective anomaly. Point anomalies can be effectively detected by reconstruction-based models (showing in Figure \ref{fig:intro1}), since they use a point-wise reconstruction error. However, contextual anomalies and collective anomalies can be hard to captured by them because they violate the normal patterns only in the ``context'' rather than obviously deviating the normal value range (anomalies in Figure \ref{fig:intro2} are hard to be identified by reconstruction error). 

3) \textit{Unexpected generalization on abnormal inputs.} The autoencoder can reconstruct anomalous samples so well because of the powerful generalization capabilities of neural networks. Many reasons can contribute to the phenomenon, such as an over-fitting autoencoder, structural similarities between anomalous and normal samples, leading an unclear boundary between anomalies and normalities. Besides, unlabeled anomalies in the training data (termed Anomaly Contamination) can significantly aggravate this issue since autoencoders are trained to minimize reconstruction error for them as well.

To address the above challenges, we propose MemAAE (\textit{Memory-augmented Adversarial Autoencoders with Deep Reconstruction and Prediction}), a novel unsupervised anomaly detection model for multivariate time-series. Specifically, we improve conventional autoencoders with three dedicated modules to solve the issues respectively. The first is the use of adversarial training \cite{goodfellow2014generative}. Unlike relevant researches  that directly use the GAN model \cite{Zenati2018AdversariallyLA,li2019mad}, we use adversarial autoencoders (AAE) \cite{makhzani2015adversarial} combined with the point-wise reconstruction objective instead. The second is an additional prediction branch. Compared with reconstruction, predicting near future values can be sensitive for contextual and collective anomalies since the prediction task incorporates temporal information over instances, which is ignored by the reconstruction task. The third is a memory mechanism. Instead of allowing the latent variable spreads the whole space or just applying some sparse regularization (such as standard Gaussian distribution), we rebuild the latent vector by the linear combination of normal patterns from the memory module. The memory module can be viewed as a dictionary that stores the basis of the normal manifold. Then the encoder plays the role of determining weights of memory slots (vectors in the memory module). In this way, the burden of modelling the normal patterns is designated to the memory module.

In summary, this paper makes the following three major contributions:

\begin{itemize}
    \item We propose a novel unsupervised framework, MemAAE, for multivariate time-series anomaly detection. It adopts an simple and effective autoencoder architecture, incorporating with an adversarial training mechanism.
    \item We extend MemAAE with a memory module, which explicitly models the normality manifold. By enforcing reconstructions are computed by only normal patterns of the memory module, MemAAE can largely avoid unexpected generalization for anomalous samples.
	\item We augment MemAAE with a prediction module, a complementary training task for the reconstruction. Forecasting future can be more sensitive for contextual and collective anomalies, which finely makes up for the reconstruction task.
	\item Through extensive experiments, we show that MemAAE achieves an \textbf{overall F1 score of 0.90} on four real-world datasets, showing the benefits of utilizing the memory module and the prediction task. Besides, further experiments including sensitivity test of hyperparameters and ablation study demonstrate effectiveness and robustness of our proposed method.
\end{itemize}

\section{Related Works}

Time-series anomaly detection is an active research topic that has been largely studied \cite{10.5555/3086742,chandola2009anomaly,chalapathy2019deep,gupta2013outlier,braei2020anomaly}. Traditional approaches of anomaly detection mainly focus on utilizing prior assumptions on normal patterns which can be categorized into distance-based \cite{ramaswamy2000efficient,chaovalitwongse2007time}, density-based \cite{breunig2000lof,ma2003time}, isolation-based \cite{liu2008isolation} and statistic-based \cite{DBLP:conf/kdd/SifferFTL17,li2020copod} methods. 

\subsubsection{Prediction-based Time-series Anomaly Detection.} Recently, deep learning empowered anomaly detection has draw extensive attention \cite{pang2021deep}. A line of works focus on utilizing prediction errors to detect anomalies since they are relatively unpredictable compared with normal points. EGADS \cite{laptev2015generic} is a generic prediction-based anomaly detection framework developed by \textit{Yahoo}. In \cite{hundman2018detecting} and \cite{filonov2016multivariate}, Long-short Term Memory (LSTM) is applied to detect anomalies via prediction errors. However, the performance of prediction-based methods heavily rely on the capability of the prediction model. Since complicated time-series are hard to be accurately predicted, the effectiveness of prediction-based anomaly detection is largely limited.

\subsubsection{Reconstruction-based Time-series Anomaly Detection.} Another paradigm of anomaly detection uses reconstruction errors to detect anomalies. To achieve this, various autoencoders are used, such as VAE \cite{xu2018unsupervised,DBLP:conf/ipccc/LiCP18}, Recurrent Neural Network (RNN) based autoencoders \cite{malhotra2016lstmbased,park2018multimodal,su2019robust,ijcai2019-378,sisvae9064715} and GAN \cite{goodfellow2014generative} based Autoencoders \cite{li2019mad,audibert2020usad}. However, reconstruction-based anomaly detection is problematic for two main reasons. Firstly, more powerful the autoencoder is, more risky that anomalies to be well reconstructed during inference phase. Well-reconstructed anomalies can heavily hurt the performance of anomaly detection. Secondly, global anomalies can be effectively captured by reconstruction errors but local contextual anomalies can be insensitive for them.

\subsubsection{Multi-task Learning for Anomaly Detection.} Multi-task learning \cite{zhang2021survey} demonstrates the effectiveness of training multiple tasks using a shared network architecture. Existed works considering multi-task learning for anomaly detection include detecting abnormal events in videos \cite{georgescu2021anomaly,zhao2017spatio} and identifying anomalous driving \cite{sadhu2019deep}. However, the application on time-series anomaly detection is largely lagging behind.

\subsubsection{Memory Models.} In the literatures, memory plays a fundamental role of building deep learning systems for both earlier and recent models, e.g. caching previous states in LSTM and Gated Recurrent Units (GRU) \cite{hochreiter1997long,chung2014empirical}, and storing history representations in contrastive representation learning \cite{wu2018unsupervised,he2020momentum}. As for anomaly detection, previous works \cite{gong2019memorizing,park2020learning} have utilized memory to prevent unexpected generalization on anomalous inputs. However, they are not prepared for time-seires and can not address all major challenges of time-series anomaly detection mentioned before.



\section{Methodology}

\begin{figure*}[ht]
	\centering
	\includegraphics[width=0.9\textwidth]{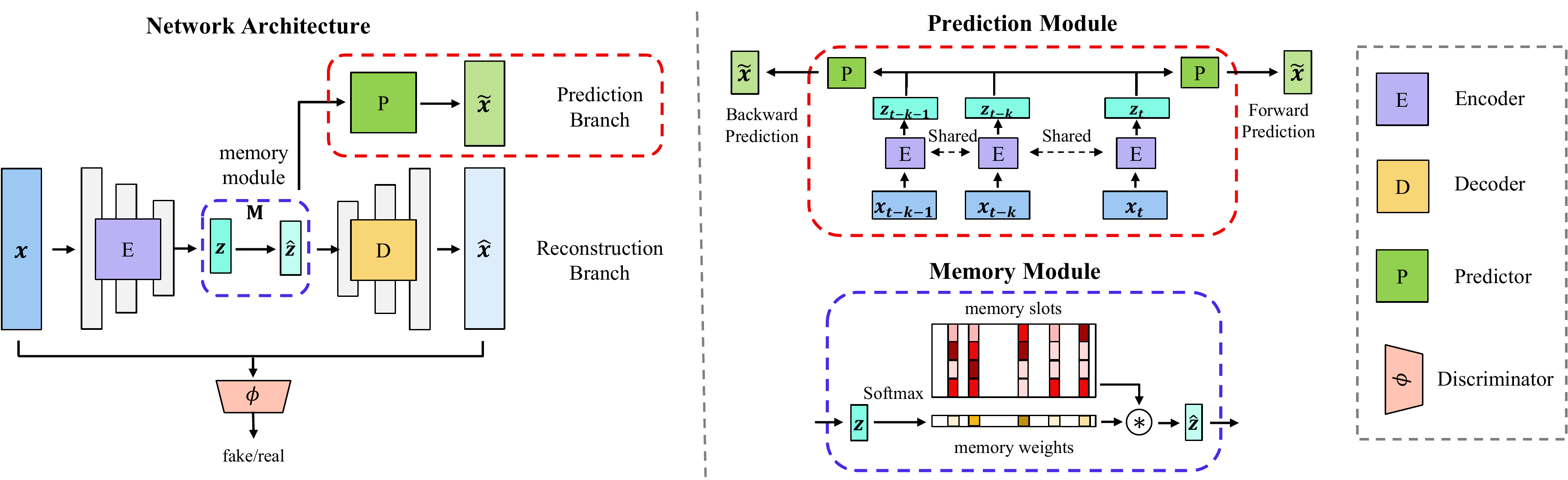}
	\caption{Systematic overview of our proposed model. MemAAE mainly contains three modules: (1) \textbf{Autoencoder with adversarial training}. This part is the basic module to perform the reconstruction task. Reconstructed samples are constrained to be as close as input samples under the ``pixel level'' (by $\ell_2$ distance) and ``global level'' (by adversarial loss) criterions simultaneously. (2) \textbf{Memory module}. As the right hand side shows, the memory module is used to model the normality manifold, which consists of a sequence of memory vectors. In the reconstruction phase, the weights of each memory vector is infered by the output of the encoder. The latent vector to be feed into the decoder is obtained by the linear combination of memory vectors over the weights. (3) \textbf{Prediction module}. To detect contextual and collective anomalies, a dedicated prediction branch is incorporated with our model. By aggreagating latent vectors over a time window, MemAAE predicts past and future values simultaneously. In the evaluation phase, reconstruction errors and prediction errors are used to detect anomalies.}
	\label{fig:framework}
\end{figure*}

\subsection{Problem Setup}

\subsubsection{Time-series anomaly detection.} Giving a multivariate time series $\boldsymbol X=\{x_1,x_2,\cdots,x_N\}, x_i \in \mathbb R^K$, of $N$ observations and $K$ variables, the goal of \textbf{time series anomaly detection} is to learn a scoring function $\phi(\cdot)$ that assigns anomaly scores to each observation such that we have $\phi(x_i) > \phi(x_j)$ if $x_i$ is anomalous and $x_j$ is normal. Detailed description of input format and data preprocessing are listed in the appendix.

\subsection{Overview}

As shown in Figure \ref{fig:framework}, MemAAE mainly consists of following modules:

\begin{itemize}
    \item \textbf{Autoencoder with adversarial training}. Same as existed time-series anomaly detection works \cite{audibert2020usad,su2019robust}, we use the Autoencoder (AE) \cite{rumelhart1985learning} as our basic model to perform the reconstruction task, where anomalies are expected to obtained higher reconstruction error. However, AE, as well as it's variant VAE \cite{DBLP:journals/corr/KingmaW13}, uses a ``pixel-level'' reconstruction error, e.g. $\ell_2$ distance of AE and log-likelihood of VAE, to measure the deviation of reconstructed samples w.r.t. input ones. The ``pixel-level'' reconstruction error can be too strict and results in over-smoothed samples (see Figure \ref{fig:intro2}). To tackle this issue, we use an additional discriminator to distinguish the true input samples and reconstructed samples \cite{goodfellow2014generative}. This technique makes reduced blurriness and enriched details for the reconstruction since the discriminator focus on similarities of the whole distribution.
    \item \textbf{Memory module}. In order to reduce well-reconstructed anomalies, we use a dedicated memory module to preserve the normality manifold. Specifically, the memory module is a dictionary of vectors where each vector is a representative point of the normality manifold. In the reconstruction process, the query, produced by the encoder, is used to draw memory vectors from the memory and the latent vector is infered by the linear combination of retrieved memory vectors. This ensures reconstructions of the decoder is strictly reconstructed by normal patterns.
    \item \textbf{Prediction module}. To detect contextual and collective anomalies, we design a prediction branch based on the latent vectors obtained by the memory module that predicts near future values. The reconstruction branch and the prediction branch share the same encoder structure. We assume that anomalies can not be well predicted since they deviate the normal patterns. In the detection phase, the anomaly score is determined by reconstruction error and prediction error together.
\end{itemize}

Next, we will describe the details of each module in the following sections.

\subsection{Memory-augumented Adversarial Autoencoder}

\subsubsection{Encoder and decoder.}\label{sec:enc_dec}

The autoencoder consists of an encoder and a decoder. The encoder $f_e(\cdot)$ takes the input $\boldsymbol x$ and maps it into a latent vector $\boldsymbol z$. The decoder $f_d(\cdot)$ maps the latent vector back to the input space to obtain the reconsturcted sample $\hat{\boldsymbol x}$. Formally, the process can be defined as:

\begin{align}
    \boldsymbol z &= f_e(\boldsymbol x),\\
    \hat{\boldsymbol x} &= f_d(\hat{\boldsymbol z}).
\end{align}

For standard autoencoders, we have $\hat{\boldsymbol z} = \boldsymbol z$. In MemAAE, we use $\boldsymbol z$ to retrieve the latent vector $\hat{\boldsymbol z}$ from memory, which will be introduced later. The deviation between the input sample and the reconstruction is measured by the reconstruction error, e.g. $\ell_2$ distance:

\begin{equation}
    \mathcal L_\text{rec} = \parallel \boldsymbol x - \hat{\boldsymbol x} \parallel_2.
\end{equation}

The objective of a standard AE is to minimize the reconstruction error $\mathcal L_\text{rec}$. Autoencoder-based anomaly detection methods use $\mathcal L_\text{rec}$ as the anomaly score while samples of high reconstruction error are considered to be anomalies.

\subsubsection{Adversarial traininig}

A common issue of the vallina autoencoder is that MSE loss leads over-smoothed reconstructions. We use adversarial training to tackle this problem. In detail, an additional discriminator $\phi(\cdot)$ is used to distinguish input samples and reconstructions. The objective of adversarial training is:

\begin{equation}
    \mathcal L_\text{adv} = \mathbb E_{\boldsymbol x\sim p_\text{data}(\boldsymbol x)} \left[\log (\phi(\boldsymbol x))\right] + \mathbb E_{\hat{\boldsymbol x} \sim p_\text{gen}(\hat{\boldsymbol x})} \left[\log (1 - \phi(\hat{\boldsymbol x}))\right],
\end{equation}
where $p_\text{data}$ denotes the ground truth data distribution and $p_\text{gen}$ denotes the distribution of reconstructed samples. Thus the overall objective of autoencoder training becomes:

\begin{equation}
    \mathcal L_\text{full-adv} = \mathcal L_\text{adv} + \lambda \cdot \mathcal L_\text{rec},
\end{equation}
where $\lambda$ is the weight of the reconstruction objective.

\subsubsection{Memory module.}\label{sec:mem}

In order to enable normality manifold preserving, we augment the autoencoder with a dedicated memory module. The latent vector to be feed into the decoder is computed by retrieving items of the memory module and taking linear combination of them. The vector produced by the encoder plays the role to compute the weights over memory slots, which means the burden of modeling normal patterns is allocated to the memory module rather than the encoder. The memory module is shared during the whole training phase. 

In detail, the memory module is composed of a matrix $\mathbf M = \left[\boldsymbol m_1, \cdots, \boldsymbol m_M\right] \in \mathbb R^{M\times K}$ where $M$ is the memory size and $K$ is the dimension of each memory slot $\boldsymbol m_i$. Given a query $\boldsymbol z$, the latent vector $\hat{\boldsymbol z}$ is computed by the linear combination of memory slots:

\begin{align}
    \hat{\boldsymbol z} = \sum_{i=1}^M \boldsymbol w_i \cdot \boldsymbol m_i = \boldsymbol w \mathbf M,
\end{align}
where $\boldsymbol w_i$ refers to the contribution of the $i$-th memory slot $\boldsymbol m_i$ and is obtained by the query $\boldsymbol z$. In practice, the query $\boldsymbol z$ is the vector produced by the encoder, namely $\boldsymbol z = f_e(\boldsymbol x)$.

The memory weights $\boldsymbol w$ over memory slots can be obtained by multiple approaches. \citet{gong2019memorizing} computes the weights via an attention manner. For simplicity, we directly assign the normalized query vector as the memory weights. Since the dimension of the query and the memory size are not required to be matched, we first map the query into the $\mathbb R^M$ space vis a projection head $\psi(\cdot)$. Then the weights are normalized by \texttt{Softmax} function:

\begin{equation}
    \boldsymbol w = \texttt{Softmax}(\psi(\boldsymbol z)).
\end{equation}



In the reconstruction task, the decoder is restricted to use representations acquired from the normality manifold only. To get reliable reconstructions, the memory module is forced to record most representative normal patterns.




\subsection{Future Forecasting for Anomaly Detection}\label{sec:pred}

The above mentioned memory-augmented reconstruction is strong enough to detect point anomalies. However, reconstruction based anomaly detection can be insensitive with contextual and collective anomalies. These anomalies are hard to be captured by reconstruction error because they deviate normal characteristics only (e.g. seasonality) or depend on previous observations. Thus we propose to leverage future prediction for anomaly detection. Compared with reconstruction, future prediction is more sensitive for local and contextual anomalies since they are hard to be predicted. The details of the future prediction for anomaly detection are introduced in the following.

\subsubsection{Structure of prediction branch.}

The prediction branch $g(\cdot)$ can be implemented by any autoregressive model, such as recurrent neural networks (RNN). In this paper, we use LSTM \cite{hochreiter1997long} to perform the prediction task. The prediction of a sample at time $t$ is computed by aggregating past $k$ latent vectors:

\begin{equation}
    \widetilde{\boldsymbol x}_t = g(\hat{\boldsymbol z}_{t-k-1}, \cdots, \hat{\boldsymbol z}_{t-1}).
\end{equation}

\subsubsection{Weighted decay loss for prediction.}
To make prediction close to the ground truth, we minimize the $\ell_2$ distance between a predicted sample $\widetilde{\boldsymbol x}_t$ at time $t$ and its ground truth $\boldsymbol x_t$:

\begin{equation}
    d(\boldsymbol x_t, \widetilde{\boldsymbol x}_t) = \parallel \boldsymbol x_t - \widetilde{\boldsymbol x}_t\parallel_2.
\end{equation}

Besides, similar with \citet{zhao2017spatio}, we assume predictions for closer time-steps are more important. Thus we reweight the contribution of each prediction according to their temporal distances. The prediction objective can be formulated as a weighted sum of predictions over predicted time-steps:
\begin{equation}
    \mathcal L_\text{pred} = \frac{1}{T^2}\sum_{i=1}^T (T-i)\cdot d(\boldsymbol x_{t+i}, \widetilde{\boldsymbol x}_{t+i}),
\end{equation}
where $T$ is the number of prediction steps.

\subsubsection{Bi-directional prediction.}
Although forecasting future is natural from the perspective of human perception, predicting backwards is also feasible for our model. Besides, leveraging bi-directional prediction for anomaly detection can make full use of data since temporal correlations are naturally bi-directional. Similar with forward prediction, we use an indentical predictor $g^\prime(\cdot)$ to perform the backward prediction:

\begin{equation}
    \widetilde{\boldsymbol x}_t = g^\prime(\hat{\boldsymbol z}_{t+k+1}, \cdots, \hat{\boldsymbol z}_{t+1}).
\end{equation}

Given the time window $\{\boldsymbol x_{t-W+1},\cdots,\boldsymbol x_t\}$, the backward prediction objective is formalized as forcasting $T$ predecessors of the window:
\begin{equation}
    \mathcal L_\text{pred-back} = \frac{1}{T^2}\sum_{i=1}^T (T-i)\cdot d(\boldsymbol x_{t-W-i}, \widetilde{\boldsymbol x}_{t-W-i}),
\end{equation}
where $W$ represents the size of the time window and $T$ indicates the number of prediction steps.

Finally, the full objective is a combination of adversarial loss, reconstruction loss and two prediction losses:

\begin{equation}\label{eq:full}
    \mathcal L_\text{full} = \mathcal L_\text{adv} + \lambda\cdot\mathcal L_\text{rec} + \gamma_1\cdot\mathcal L_\text{pred} + \gamma_2\cdot \mathcal L_\text{pred-back},
\end{equation}
where $\lambda$, $\gamma_1$ and $\gamma_2$ are the weights of reconstruction loss, forward prediction loss and backward prediction loss respectively.

\subsection{Anomaly Detection}

In the detection stage, we calculate anomaly scores for testing samples based on our trained model. As described in the above sections, anomalies are expected to obtain larger reconstruction error and prediction error. Thus we use them to calculate the anomaly score. Formally, the anomaly score $\mathcal S(\cdot)$ of a sample $\boldsymbol x$ is defined as :

\begin{equation}
    \mathcal S(\boldsymbol x) = \lambda\cdot\mathcal L_\text{rec} + \gamma_1\cdot\mathcal L_\text{pred} + \gamma_2\cdot \mathcal L_\text{pred-back},
\end{equation}
where $\lambda$, $\gamma_1$ and $\gamma_2$ are weights to adjust each term, with same values in Equation \ref{eq:full}.

In real-world applications, weather $\boldsymbol x$ is an anomaly can be identified by a pre-defined threshold, which is choosed by the operator. Following recent researches \cite{xu2018unsupervised,DBLP:conf/ipccc/LiCP18,audibert2020usad}, we focus on anomaly scoring rather than threshold defining.

\section{Experiments}

\subsection{Datasets and Evaluation Metrics}

We use four public multivariate time-series datasets to evaluate our model: \textit{Soil Moisture Active Passive (SMAP) satellite, Mars Science Laboratory (MSL) rover Datasets} \cite{hundman2018detecting}, \textit{Server Machine Dataset (SMD)} \cite{su2019robust} and \textit{Secure Water Treatment (SWaT) Dataset} \cite{goh2016dataset}. Detailed description of datasets is listed in the appendix.

Following previous researches \cite{audibert2020usad,su2019robust}, we use F1 score (F1) to evaluate the performance of anomaly detection, with Precision (P) and Recall (R) for reference. In the practical situation, the importance of precision and recall is varied according to different practical applications. For example, people often require higher recall to avoid missing fraud events in fraud detection while the requirement for precision is higher than recall in the field of operation and maintenance. Thus, we mainly use the F1 score to evaluate the performance to remove the preference of precision and recall. Besides, in order to remove the influence of threshold selection, we use brute force searching to get the best F1 score. Precision and Recall are calculated using the threshold of the best F1 score. Finally, in real-world applications, the point-wise metrics are not necessary for human operators. Alerting for any point in a contiguous anomaly segment is acceptable if the delay is not out of tolerance. We follow the settings of the modified F1 score which is widely accepted in literature \cite{audibert2020usad,su2019robust}. In detail, for an anomaly segment, as long as a certain point in the abnormal segment is detected by the anomaly detection method, the whole segment is considered to be abnormal. 

Finally, all experiments are conducted at a node in a Xeon cluster with 32GB dedicated memory, 4 CPU cores and 1 Tesla K80 or Geforce GTX 2080 GPU accelerator.

\begin{table*}[ht]
\centering
\caption{Performance of MemAAE and baselines.}
\footnotesize
\setlength{\tabcolsep}{0.8mm}{
\begin{tabular}{@{}ccccccccccccc@{}}
\toprule
\multirow{2}{*}{\textbf{Model}} & \multicolumn{3}{c}{\textbf{SMAP}} & \multicolumn{3}{c}{\textbf{MSL}} & \multicolumn{3}{c}{\textbf{SMD}} & \multicolumn{3}{c}{\textbf{SWaT}} \\ \cmidrule(l){2-13} 
                                & P         & R         & F1        & P         & R         & F1       & P         & R         & F1       & P         & R         & F1        \\ \midrule
IF                              & 0.4423    & 0.5105    & 0.4671    & 0.5681    & 0.6740    & 0.5984   & 0.5938    & 0.8532    & 0.5866   & 0.9620    & 0.7315    & 0.8311    \\ \midrule
DAGMM                           & 0.6334    & 0.9984    & 0.7124    & 0.7562    & 0.9803    & 0.8112   & 0.6730    & 0.8450    & 0.7231   & 0.8292    & 0.7674    & 0.7971    \\ \midrule
AE                              & 0.7216    & 0.9795    & 0.7776    & 0.8535    & 0.9748    & 0.8792   & 0.8825    & 0.8037    & 0.8280   & 0.9913    & 0.7040    & 0.8233    \\ \midrule
LSTM-VAE                        & 0.7164    & 0.9875    & 0.7555    & 0.8599    & 0.9756    & 0.8537   & 0.8698    & 0.7879    & 0.8083   & 0.7123    & 0.9258    & 0.8051    \\ \midrule
OmniAnomaly                     & 0.7585    & 0.9756    & 0.8054    & 0.9140    & 0.8891    & 0.8952   & 0.9809    & 0.9438    & 0.9441   & 0.7223    & 0.9832    & 0.8328    \\ \midrule
USAD                            & 0.7697    & 0.9831    & 0.8186    & 0.8810    & 0.9786    & 0.9109   & 0.9314    & 0.9617    & 0.9382   & 0.9870    & 0.7402    & 0.8460    \\ \midrule
MemAAE                          & 0.8111    & 0.9044    & \textbf{0.8552}    & 0.9112    & 0.9298    & \textbf{0.9204}   & 0.9539    & 0.9439    & \textbf{0.9489}   & 0.9551    & 0.8239    & \textbf{0.8847}    \\ \bottomrule
\end{tabular}
}
\label{tab:comp_sota}
\end{table*}

\subsection{Comparison with SOTAs}

This section examines the effectiveness of MemAAE on real-life anomaly detection datasets with state-of-the-art competing methods.

\textbf{Setup.} We compare our proposed model with six state-of-the-art unsupervised anomaly detection methods, including Isolation Forests (IF) \cite{liu2008isolation}, Deep Autoencoding Gaussian Mixture Model (DAGMM) \cite{zong2018deep}, AE, LSTM-VAE \cite{park2018multimodal}, OmniAnomaly \cite{su2019robust} and USAD \cite{audibert2020usad}. In the experiment, Precision, Recall and adjusted F1 score are reported for each dataset.

\textbf{Results \& Analysis. } Table \ref{tab:comp_sota} shows the experimental results of our approach and baselines on the four datasets. As shown in the table, MemAAE shows superior performance and achieves best F1 scores, with gains of $1\%$, $4\%$, $0.5\%$ and $4\%$ on MSL, SMAP, SMD and SWAT than the best baseline respectively. The experimental results demonstrate the effectiveness of our proposed method. \textbf{It is noticed that precision and recall scores are calculated using the best F1 score's threshold. Thus they only imply that, under the best F1 situation, how are the precision and recall scores.} The experimental results demonstrate our method achieves the best trade-off between precision and recall. 

In detail, IF presents the lowest performance among all baselines. The limitation of IF lies in lacking the ablity to model complex normal patterns since it simply assumes that anomalies are more likely to be linearly separated in the original space. MemAAE leverages deep autoencoding to detect anomalies, which enables our method to outperform IF.

DAGMM obtains relative poor results on the four datasets. Although it utilizes deep neural networks, DAGMM focuses on anomaly detection for multivariate features without temporal correlations. It fails to model temporally dependent observations thus the performance is lagging behind. In our model, the temporal information is explicitly modeled by the prediction task. As a result, MemAAE outperforms DAGMM on all the four datasets.

Rather than conventional AE, LSTM-VAE simply stacks LSTM and VAE to model the temporal correlations. However, this approach directly replaces the feed-forward network in VAE with LSTM, i.e. aggregating information from the original inputs. From the perspective of dynamic systems, it is beneficial to summarize information from latent stochastic variables, e.g. latent vectors \cite{su2019robust}. This explains the worse performance of LSTM-VAE w.r.t models using stochastic recurrent networks such as OmniAnomaly.

OmniAnomaly models temporal correlations by stochastic recurrent networks plus a normalizing flow module to deal with complex latent distributions. However, compared with MemAAE, it is still lagged behind. The main drawback of OmniAnomaly are two-fold. Fist, the use of VAE tends to produce over-smoothed reconstructions, which ignores slight perturbations of anomalies. Second, OmniAnomaly only uses reconstruction errors to detect anomalies, which may neglects local and contextual anomalies. As a result, MemAAE still outperforms it on all datasets.

USAD incorporates the autoencoder with adversarial training. Compared with OmniAnomaly and other reconstruction-based approaches, USAD uses an additional discriminator to amplify ``mild'' anomalies, which can be more sensitive for slight anomalies. However, MemAAE still exceeds USAD on all datasets. The main difficulty of USAD is that it suffers the burden of training two contronting discriminators. As demonstrated in prior works \cite{DBLP:conf/icml/ArjovskyCB17}, GAN is notoriously hard to be trained. USAD may sometimes converge to a sub-optimal equilibrium, which can heavily hurt the performance.

\begin{figure*}[h]
\centering
\begin{subfigure}{0.25\textwidth}
	\centering
	\includegraphics[width=\linewidth]{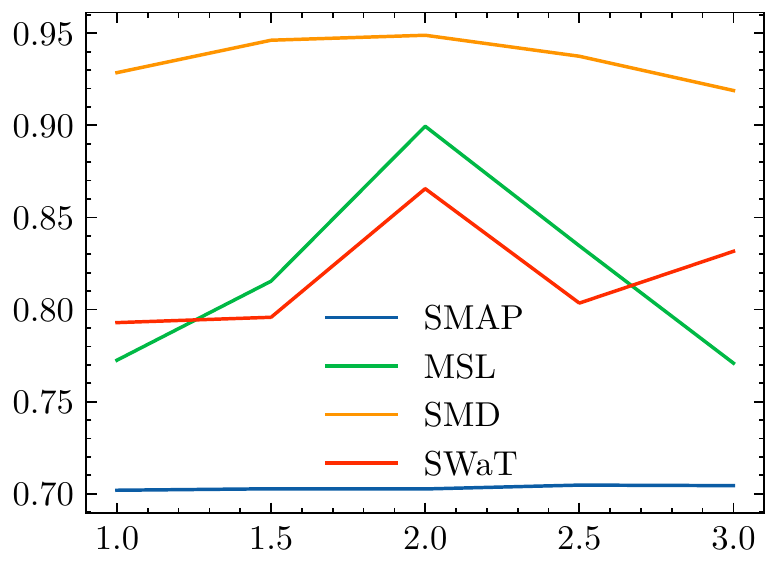}
	\caption{Reconstruction weights.}
	\label{fig:rec_weight}
\end{subfigure}
\begin{subfigure}{0.25\textwidth}
	\centering
	\includegraphics[width=\linewidth]{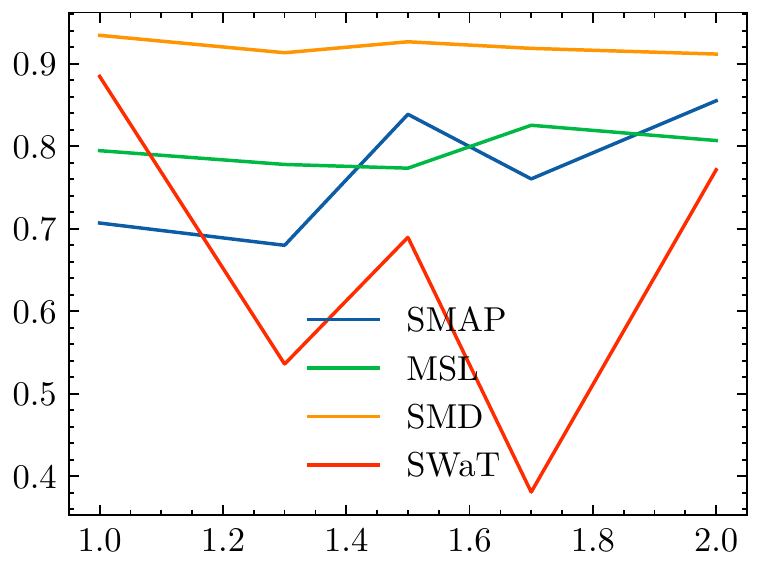}
	\caption{Forward prediction weights.}
	\label{fig:forward_pred}
\end{subfigure}
\begin{subfigure}{0.25\textwidth}
	\centering
	\includegraphics[width=\linewidth]{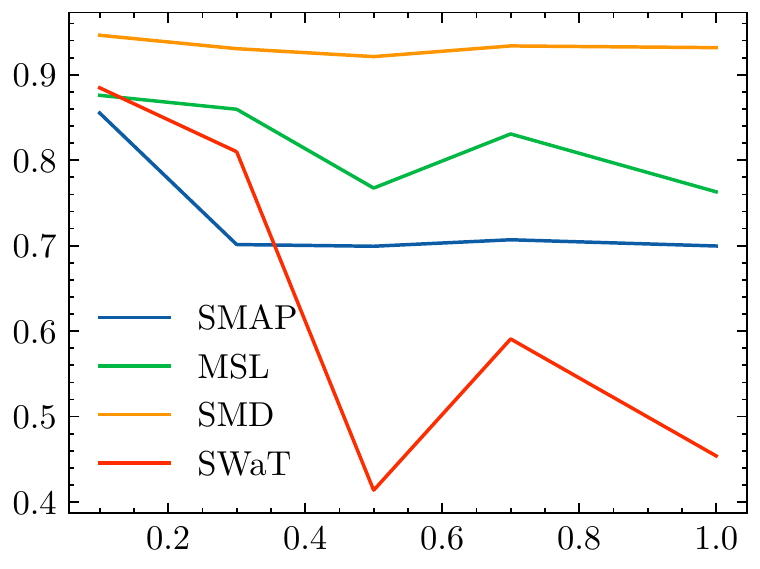}
	\caption{Backward prediction weights.}
	\label{fig:back_pred}
\end{subfigure}
\caption{Sensitivities test for reconstruction weight $\lambda$, forward prediction weight $\gamma_1$ and backward prediction weight $\gamma_2$ respectively.}
\label{fig:sensitivity}
\end{figure*}

\subsection{Sensitivities w.r.t. Hyperparameters}\label{sec:sensitivity_hyperparams}

\textbf{Setup.} In this section, we study the effects of different hyperparameters to the performance of MemAAE, including the reconstruction weight, the prediction weights, the size of the memory module, and the number of prediction steps. We adjust the setting of each parameter respectively and report the performance for all the four datasets.

\begin{table}[h]
\centering
\caption{Quantitative results of different prediction steps.}
\scriptsize
\setlength{\tabcolsep}{0.4mm}{
\begin{tabular}{@{}ccccccccccccc@{}}
\toprule
\multirow{2}{*}{\textbf{Pred Step}} & \multicolumn{3}{c}{\textbf{SMAP}} & \multicolumn{3}{c}{\textbf{MSL}} & \multicolumn{3}{c}{\textbf{SMD}} & \multicolumn{3}{c}{\textbf{SWaT}} \\ \cmidrule(l){2-13} 
                                    & P         & R         & F1        & P         & R         & F1       & P         & R         & F1       & P         & R         & F1        \\ \midrule
5                                   & 0.977    & 0.552    & 0.705    & 0.842    & 0.795    & 0.818   & 0.878    & 0.921    & 0.899   & 0.557    & 0.733    & 0.633    \\ \midrule
10                                  & 0.955    & 0.562    & 0.707    & 0.820    & 0.924    & 0.869   & 0.955    & 0.885    & 0.919   & 0.952    & 0.715    & 0.817    \\ \midrule
15                                  & 0.789    & 0.722    & 0.754    & 0.809    & 0.786    & 0.797   & 0.929    & 0.931    & 0.930   & 0.585    & 0.715    & 0.643   \\ \midrule
20                                  & 0.933    & 0.564    & 0.703    & 0.825    & 0.781    & 0.802   & 0.934    & 0.919    & 0.926   & 0.993    & 0.707    & 0.826    \\ \bottomrule
\end{tabular}
\label{tab:pred_step}
}
\end{table}

\begin{table}[h]
\centering
\caption{Quantitative results of different memory sizes.}
\scriptsize
\setlength{\tabcolsep}{0.4mm}{
\begin{tabular}{@{}ccccccccccccc@{}}
\toprule
\multirow{2}{*}{\textbf{Mem Size}} & \multicolumn{3}{c}{\textbf{SMAP}} & \multicolumn{3}{c}{\textbf{MSL}} & \multicolumn{3}{c}{\textbf{SMD}} & \multicolumn{3}{c}{\textbf{SWaT}} \\ \cmidrule(l){2-13} 
                                   & P         & R         & F1        & P         & R         & F1       & P         & R         & F1       & P         & R         & F1        \\ \midrule
128                                & 0.977    & 0.550    & 0.703    & 0.935    & 0.880    & 0.906   & 0.935    & 0.918    & 0.926   & 0.740    & 0.754    & 0.747    \\ \midrule
256                                & 0.972    & 0.552    & 0.704    & 0.774    & 0.883    & 0.825   & 0.934    & 0.906    & 0.920   & 0.878    & 0.723    & 0.793    \\ \midrule
512                                & 0.982    & 0.549    & 0.704    & 0.884    & 0.869    & 0.872   & 0.878    & 0.921    & 0.899   & 0.940    & 0.706    & 0.806    \\ \midrule
1024                               & 0.943    & 0.560    & 0.703    & 0.939    & 0.801    & 0.865   & 0.942    & 0.920    & 0.931   & 0.427    & 0.732    & 0.540    \\ \bottomrule
\end{tabular}
\label{tab:mem_size}
}
\end{table}

\textbf{Reconstruction Weight $\boldsymbol\lambda$.} Firstly, we study the parameter of reconstruction weight. A larger reconstruction weight indicates more attention to the reconstruction loss contained in the overall objective. Figure \ref{fig:rec_weight} shows the experimental results of different reconstruction weights, in which range from $[1.0, 1.5 , 2.0, 2.5, 3.0]$. From the experimental results, we can find that on SMAP and SMD, the performance of MemAAE is relatively stable w.r.t. reconstruction weights. Besides, the change of reconstruction weights causes large fluctuations to the performance of MemAAE on MSL and SWaT, indicating that the reconstruction loss plays a crucial role in the determination of anomalies on MSL and SWAT.

\textbf{Prediction Weights $\boldsymbol\gamma_1$ and $\boldsymbol\gamma_2$.} We investigate the influence of two prediction weights $\gamma_1$ and $\gamma_2$. As described above, the prediction task is designed to detect local anomalies. Since the significance of local anomalies is varied over different datasets, the choice of the prediction weight is of great importance. To study the influence of them, we use candidate settings of $[1.0, 1.3, 1.5, 1.7, 2.0]$ and $[0.1, 0.3, 0.5, 0.7, 1.0]$ for the forward prediction weight and the backward prediction weight respectively. The experimental results of the two parameters are shown in Figure \ref{fig:forward_pred} and Figure \ref{fig:back_pred} separately. We can investigate that MSL and SMD are relatively stable for different parameter settings. Besides, SMAP and SWaT are extremely sensitive for the prediction weight. According to the experiments, we choose the settings of prediction weights empirically.

\textbf{Prediction Steps $\boldsymbol T$.} Here we study the influence of the number of prediction steps. The prediction module performs a bi-directional forecasting to detect anomalies, especially local anomalies. As shown in Table \ref{tab:pred_step}, extending the range of forecasting can enlarge the receptive field of the predictor but it can not always bring a performance boost. We can investigate that the best setting of prediction steps largely depends on the specific dataset. 

\textbf{Memory Size $\boldsymbol M$.} Lastly, we study the influence of the memory size. The memory module preserves normal patterns to suppress the generalization capabilities of the autoencoder. By reconstructing samples only from the memory module, anomalies are less likely to be well reconstructed. As illustrated in Table \ref{tab:mem_size}, the choice of the memory size is important. Simply enlarging the scale of the memory can obtain insignificant gains (SMAP), even a performance drop (SWaT). The experimental results suggest us to select a proper value for the memory size.

\subsection{Ablation Study}

\textbf{Setup.} In this section, we examine the effectiveness of our proposed two modules: the prediction module and the memory module. In the experiment, we compare our model with its simplified counterparts, denoted as \textit{w/o prediction} and \textit{w/o memory}, respectively.

\begin{table}[h]
\centering
\caption{Quantitative results of MemAAE and its variants.}
\scriptsize
\setlength{\tabcolsep}{0.4mm}{
\begin{tabular}{@{}ccccccccccccc@{}}
\toprule
\multirow{2}{*}{\textbf{Model}} & \multicolumn{3}{c}{\textbf{SMAP}} & \multicolumn{3}{c}{\textbf{MSL}} & \multicolumn{3}{c}{\textbf{SMD}} & \multicolumn{3}{c}{\textbf{SWaT}} \\ \cmidrule(l){2-13} 
                                & P         & R         & F1        & P         & R         & F1       & P         & R         & F1       & P         & R         & F1        \\ \midrule
\textit{w/o prediction}                  & 0.939    & 0.566    & 0.706    & 0.844    & 0.725    & 0.780   & 0.918    & 0.945    & 0.931   & 0.974    & 0.719    & 0.827    \\ \midrule
\textit{w/o memory}                      & 0.974    & 0.548    & 0.702    & 0.846    & 0.830    & 0.838   & 0.922    & 0.909    & 0.915   & 0.972    & 0.731    & 0.834    \\  \midrule
MemAAE                          & 0.811    & 0.904    & 0.855    & 0.911    & 0.930    & 0.920   & 0.954    & 0.944    & 0.949   & 0.955    & 0.824    & 0.885    \\ \bottomrule
\end{tabular}
}
\label{tab:ablation}
\end{table}

\textbf{Results \& Analysis.} As illustrated in Table \ref{tab:ablation}, the memory module and the prediction module all have major contributions to the superior performance of MemAAE. Removing any of them results in degrading the original performance of the model significantly. 

The memory module enables MemAAE to reduce false negatives, i.e. well-reconstructed anomalies in the evaluation phase. Without the memory module, the model can ``generalize'' so well such that it also reconstruct anomalies well, which will degenerate the performance. Besides, the anomalies can also appeared in the training dataset (termed \textit{Anomaly Contamination}), since we follow a fully unsupervised setting. In this situation, the encoded latent space can be polluted by anomalous patterns without using the memory module. Thus, the memory module plays a vital role to avoid this issue.

The prediction module provides a complementary anomaly scoring approach for reconstruction errors. Generally, the reconstruction task is good at detecting global anomalies, but it may ignore local perturbations or contextual deviations, especially when the values still conform to normal range. With the help of the prediction task, this situation can be tackled since these incident deviations can harded to be predicted. By incorporating the reconstruction task, MemAAE can detect a wide range of anomalies, leading robust and effectives anomaly detection in real-world applications.

\section{Conclusion}

In this paper, we propose a novel unsupervised algorithm, MemAAE, for multi-variate time series anomaly detection. Main novelties of this work are three-fold. First, by taking the advantage of adversarial training, MemAAE is able to reconstruct complex normal patterns of time-series. Second, the memory module encodes the normality manifold into memory vectors. By composing vectors drawn from the memory module, reconstructions are only computed from normal patterns. Finally, the prediction branch plays a complementary role for the reconstruction branch since forecasting future can be sensitive for local anomalies, which are hard to be captured by the reconstruction task. Extensive experiments with real-world datasets show the effectiveness and robustness of our proposed method compared with competing baselines.


\bibliography{aaai22}

\clearpage
\appendix
\section{Appendix}

\subsection{Data Preprocessing}

\subsubsection{Data normalization.} We rescale the time-series into the range $[0, 1]$ using minimum values and maximum values of the training data:

\begin{equation}
    \widetilde{x} = \frac{x - \min(X_\text{train})}{\max(X_\text{train}) - \min(X_\text{train})},
\end{equation}
where $\max(X_\text{train})$ and $\min(X_\text{train})$ denote the maximum value and the minimum value of training data respectively.

\subsubsection{Sliding window.} Instead of feeding observations at each timestamp to the model, we consider the observation and it's context predecessors. Thus, we apply sliding window with a length $W\ll N$ over the time-series and get the reformed dataset $\mathbf X\in\mathbb R^{(N-W+1) \times W \times K}$ where each window $\boldsymbol x \in \mathbb R^{W\times K}$ is the basic element.

\subsection{Implementation Details}

In terms of network architectures, we implement the autoencoder via one dimensionl convolutional networks (1D ConvNet). We choose convolutional networks rather than fully connected networks, e.g. \cite{xu2018unsupervised,su2019robust,audibert2020usad}, because they are more computationally effective for large datasets. The discriminator shares the same architecture with the encoder, but outputs a single real value. The prediction module is composed of bi-directional LSTM \cite{birnn650093}, with 2 stacked fully-connected layer for each recurrent cell. We use Adam \cite{DBLP:journals/corr/KingmaB14} to perform the gradient descent with a learning rate of $10^{-3}$. To guarantee a comprehensive optimization, we train our model using 150 epochs with 512 mini-batches in each epoch. Other quantitative settings of hyperparameters are described in Table \ref{tab:datasets}.

\subsection{Dataset Details}

The details of datasets are described as following:

\begin{itemize}
    \item \textit{Soil Moisture Active Passive (SMAP) satellite and Mars Science Laboratory (MSL) rover Datasets} \footnote{\url{https://github.com/khundman/telemanom}}. SMAP and MSL are released by NASA \cite{hundman2018detecting} with annotations collected from real-world. They contain the data of 55 and 27 entities with 25 and 55 metrics for each respectively.
    \item \textit{Server Machine Dataset (SMD)} \footnote{\url{https://github.com/NetManAIOps/OmniAnomaly}}. SMD \cite{su2019robust} is publicly released by a large Internet company and the data is monitored by 38 variables with time spanning of 5 weeks. It is divided into two subsets with equal size: one for training and the other for testing.
    \item \textit{Secure Water Treatment (SWaT) Dataset} \footnote{\url{https://itrust.sutd.edu.sg/testbeds/secure-water-treatment-swat/}}. SWaT \cite{goh2016dataset} is a scaled down version produced by a real-world industrial water treatment plant to understand the conditions under cyber attacks. It is systematically generated from the testbed with 11 days of continuous operation: 7 days under normal operation and 4 days with attack scenarios from all the 51 sensors and actuators. 
\end{itemize}

\begin{table}[h]
\centering
\caption{Statistics and parameter settings of each dataset.}
\footnotesize
\setlength{\tabcolsep}{1.0mm}{
\begin{tabular}{@{}lcccc@{}}
\toprule
                     & \textbf{SMAP} & \textbf{MSL} & \textbf{SMD} & \textbf{SWaT} \\ \midrule
Train                & 135,183       & 58,317       & 708,405      & 496,800       \\
Test                 & 427,617       & 73,729       & 708,420      & 449,919       \\
Dimensions           & 55$\times$25  & 27$\times$55 & 28$\times$38 & 51            \\
Anomaly ratio ($\%$) & 13.13         & 10.72        & 4.16         & 11.98         \\ \midrule
Window size          & 32            & 32           & 32           & 32             \\
Latent size          & 16            & 16           & 16           & 16              \\
Memory size          & 512           & 512          & 512          & 512             \\
Pred step            & 7             & 7          & 30          & 5             \\
Reconstruction weight           & 1.0           & 1.0          & 2.0          & 1.1              \\
Forward prediction weight   & 2.0           & 1.4          & 0.2          & 1.0              \\
Backward prediction weight  & 0.1           & 0.2          & 0.1          & 0.1              \\
    \bottomrule
\end{tabular}
}
\label{tab:datasets}
\end{table}

\end{document}